\documentclass[10pt,twocolumn,letterpaper]{article}

\usepackage[pagenumbers]{cvpr} 

\usepackage{graphicx}
\usepackage{amsmath}
\usepackage{amssymb}
\usepackage{booktabs}
\usepackage{pifont}
\usepackage{xcolor}
\usepackage{arydshln}
\newcommand{\cmark}{\textcolor{green!80!black}{\ding{51}}}
\newcommand{\xmark}{\textcolor{red}{\ding{55}}}
\usepackage{stmaryrd}

\usepackage{algorithm}
\usepackage{algpseudocode}
\usepackage{amsmath}

\usepackage{mathtools}
\usepackage[textsize=tiny]{todonotes}
\usepackage[accsupp]{axessibility}

\usepackage[pagebackref,breaklinks,colorlinks]{hyperref}

\usepackage[capitalize]{cleveref}
\crefname{section}{Sec.}{Secs.}
\Crefname{section}{Section}{Sections}
\Crefname{table}{Table}{Tables}
\crefname{table}{Tab.}{Tabs.}

\def\confName{CVPR}
\def\confYear{2023}

\begin{document}

\title{Generating Dialogues from Egocentric Instructional Videos for Task Assistance: Dataset, Method and Benchmark}
\author{Lavisha Aggarwal\\{lavishaggarwal@google.com}\\
\and
Vikas Bahirwani\\{vikasbahirwani@google.com}\\
\and
Lin Li\\{linspeaking@google.com}\\
\and
Andrea Colaco\\{andreacolaco@google.com}\\
\\
}

\maketitle


\begin{abstract}
Many everyday tasks ranging from fixing appliances, cooking recipes to car maintenance require expert knowledge,  especially when tasks are complex and multi-step. Despite growing interest in AI agents, there is a scarcity of dialogue-video datasets grounded for real world task assistance. In this paper, we propose a simple yet effective approach that transforms single-person instructional videos into task-guidance two-person dialogues, aligned with fine grained steps and video-clips. Our fully automatic approach, powered by large language models, offers an efficient alternative to the substantial cost and effort required for manual data collection. Using this technique, we build HowToDIV, a large-scale dataset containing 507 conversations, 6636 question-answer pairs and 24 hours of videoclips across diverse tasks in cooking, mechanics, and planting. Each session includes multi-turn conversation where an expert teaches a novice user how to perform a task step by step, while observing user's surrounding through a camera and microphone equipped wearable device. We establish the baseline benchmark performance on HowToDIV dataset through Gemma-3 model for future research on this new task of dialogues for procedural-task assistance. Our dataset and code are publicly available at our project page: ~\url{https://github.com/google/howtodiv}.

\end{abstract}

\section{Introduction}
\label{sec:intro}
\begin{figure}[ht]
    \centering
    \includegraphics[width=0.45\textwidth]{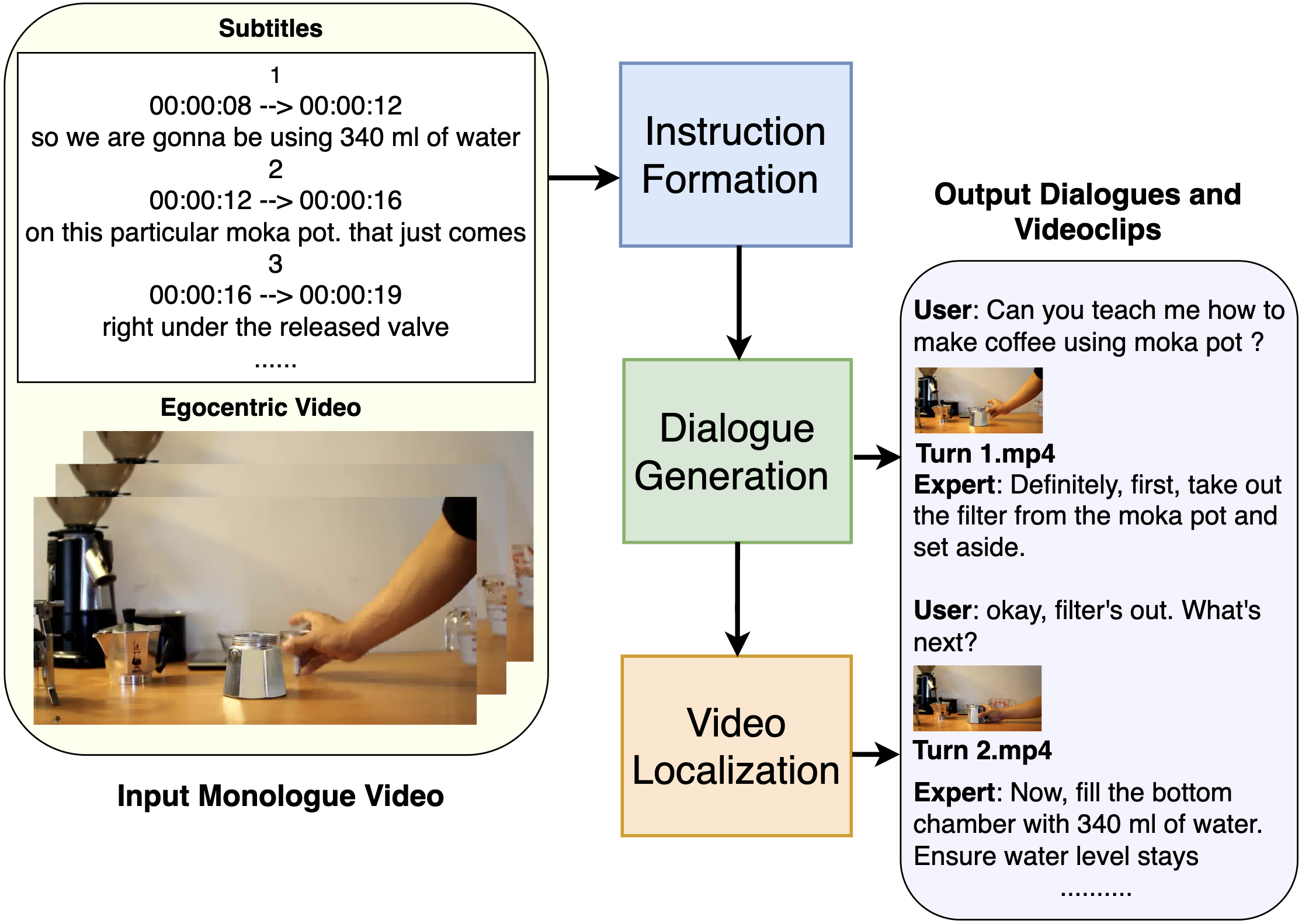}
    \caption{Overview of our approach, an automatic monologue instructional video to user-expert dialogue generation method leveraging pre-trained large language models. Starting from expert subtitles and video, we produce instructions for the given task, followed by two-person dialogues and video clips corresponding to each user turn.
    }
    \label{fig:overview}
\end{figure}
There has been a recent emergence of general-purpose AI agents assisting people and transforming every aspect of life. One such aspect involves the need for people to learn and perform new procedural tasks\footnote{\color{black} Procedural task is defined as a task consisting of multiple well-defined steps to achieve an end goal. While steps can be performed in a few valid orders, some deviations are considered  as user errors or mistakes.}, such as assembling IKEA furniture, cooking new recipes, or performing mechanical tasks like repairing home appliances, or car or bike parts. These tasks consist of multiple complex steps, and are difficult to memorize and recall precisely, owing to various nuances and details involved. The advent of smart wearable devices embedded with AI agents along with sensors such as microphones, speaker, and cameras presents us a unique opportunity to build language vision models and agents for such procedural task assistance. However the lack of datasets, catering to the domain of step by step task assistance, makes it difficult to evaluate the performance of existing models and to further improve them.

Datasets mimicking a real expert teaching a novice user how to initiate and complete a custom task can significantly aid boosting research in this direction. The user is expected to be equipped with a wearable device streaming real-time audio and video. This is fed to a vision-language model to respond to the user's questions, guide them through the different steps and help finish the task. The AI agent is expected not only to know and describe task instructions one step at a time, but also to answer intermediate questions, clarify requested details, provide feedback and be vigilant enough to check for any user errors and correct them proactively. 

HoloAssist~\cite{holoassist} is a recent egocentric dataset consisting of a continuous video stream and dialogues for collaborative object manipulation tasks built via HoloLens between two users. However, HoloAssist is restricted to 20 tasks, limiting its scope along with requiring an enormous data collection undertaking with hundreds of hours of human oversight, annotation, making it both resource-intensive and difficult to scale for diverse use cases.

Despite scanty dialogue datasets, we observed that a large number of instructional video datasets, such as NIV~\cite{niv}, EgoPER~\cite{egoper}, YouCookII~\cite{youcook2}, Assembly101~\cite{assembly101}, CrossTask~\cite{crosstask}, EpicKITCHEN~\cite{epickitchen}, COIN~\cite{coin}, are readily available. Such video datasets consist of video recordings showing an expert demonstrating how to perform a specific task while narrating the instructions for different steps. These datasets, often extracted through Youtube videos, do not contain multi-person conversations or dialogues, rather a single person delivering a monologue and are meant for post consumption instead of real-time step-by-step guidance. They are sometimes annotated with fine grained steps for the given procedure. 

In order to build a large-scale task assistance dataset, spanning tasks from various spheres of life, human-facilitated data collection is infeasible due to high cost incurred as well as the large number of human hours needed for in-field collection, supervision \& annotation. Therefore, it is of high research and production interest to explore cheaper alternatives leveraging existing models, datasets and knowledge pools.

To this end, we propose a simple yet effective prompting-based method, for generating dialogues and video-clips by fusing the language interpolation and manipulation skills of a large language model (LLM) with domain-knowledge captured in instructional videos. Our algorithm consists of three main components: instruction formation, dialogue generation and video localization, as shown in Fig~\ref{fig:overview}. 

We utilize this algorithm to generate a large-scale task-assistance dataset, HowToDIV, built on top of two popular instructional video datasets: Narrated Instruction Videos (NIV)~\cite{niv} and Egocentric Procedural Error (EgoPER)~\cite{egoper}. It consists of real human-like conversations between an expert and a novice user, along with corresponding user's point-of-view video-clips for every user-dialogue turn, and instruction steps for every conversation. The text generation is controllable with prompts and step annotations accompanying each dataset.

Given a single person video with subtitles, our approach can generate user-expert Dialogues, Instructions and corresponding user Videoclips, without any manual intervention. We benchmark the HowToDIV dataset on the Gemma3~\cite{gemma3} LLM, evaluating the metrics: BLEU score, ROUGE score and LLM as a Judge score. Gemma3 responses yield an average score of 2.87 for LLM as a Judge score (on a scale of 1 - 5), 0.32 BLEU score, 0.270 ROUGE-L score. In summary, our contributions are following:
\begin{itemize}
    \item We propose an instructional monologue video to two-person dialogue generation technique, fusing a large language model’s inherent knowledge with domain information from procedural task video datasets.
    \item We contribute HowToDIV, a large-scale multi-turn dataset consisting of 507 conversations, and 6636 user-expert dialogue turns and videos covering a diverse set of tasks across cooking, mechanical fixing and planting domains. The sessions consist of variations for user speech-styles (regular and concise), action-types (users accurately following steps vs making errors), along with the same task being performed with different step orderings.
    \item We establish the benchmark performance of Gemma3, 4B model on HowToDIV dataset, achieving a BLEU score of 0.32, ROUGE-2 score of 0.125 and LLM-as-a-Judge score of 2.97 (on a 1 - 5 scale) as the baseline for the task of step-by-step procedural-task assistance dialogues.
\end{itemize}

\section{Related Works}
\label{sec:related_works}

\textbf{Instructional Video Datasets.}
Instructional videos are a rich source for learning complex multi-step tasks, ranging from cooking recipes to car maintenance and appliance repair. Large-scale efforts like HowTo100M ~\cite{howto100m} have significantly advanced the field by collecting 136 million video clips from 1.22 million narrated instructional web videos, covering over 23,000 distinct visual tasks. This dataset leverages automatically transcribed narrations as a form of weak supervision for learning text-video embeddings, demonstrating a scalable alternative to the expensive and time-consuming manual annotation required by smaller datasets such as MSR-VTT~\cite{msrvtt}, YouCook2~\cite{youcook2}, and EPIC-KITCHENS~\cite{epickitchen}. Other notable datasets, including COIN~\cite{coin} and CrossTask~\cite{crosstask}, have also been constructed by searching instructional videos related to WikiHow~\cite{wikihow} articles, further contributing to task understanding. While these collections have been instrumental in advancing research in action recognition, segmentation, and procedural planning, they primarily consist of monologue instructions. They often lack the interactive, multi-turn dialogue crucial for real-world task assistance, a gap which our HowToDIV approach aims to address by transforming these monologues into structured conversations.

\textbf{Egocentric Video Datasets for Procedural Tasks.}
The egocentric perspective, capturing a first-person view of activities, provides unique and rich information about a user's intentions and their direct interactions with objects in the environment. This has spurred the creation of various egocentric video datasets, from general collections like Ego4D~\cite{ego4d}, spanning thousands of hours of daily life activities, to more specialized ones such as EPIC-KITCHENS~\cite{epickitchen} for culinary tasks and Assembly101~\cite{assembly101} for procedural assembly activities. More recently, datasets have emerged that focus on human-human interaction within egocentric procedural tasks. HoloAssist~\cite{holoassist} is a large-scale egocentric dataset featuring a two-person collaborative setting where a performer, wearing an AR headset, is verbally guided by a remote instructor through physical manipulation tasks. It captures comprehensive multimodal data* including RGB, depth, 3D hand pose, eye gaze, audio, and IMU, augmented with annotations for actions, mistake detection, and intervention types, highlighting the importance of proactive and grounded guidance. Similarly, EgoPER~\cite{egoper} offers an egocentric procedural error dataset for cooking, providing a detailed taxonomy of various error types (omission, addition, modification, slip, and correction) across its multimodal video content. While these datasets significantly enrich our understanding of egocentric task execution and human interaction, they do not intrinsically offer the expert-novice dialogue structures essential for dynamic, interactive AI assistants. Our HowToDIV dataset is built upon and complements such egocentric video resources (like NIV~\cite{niv} and EgoPER~\cite{egoper} datasets) by explicitly focusing on the generation of interactive dialogues.

\begin{figure*}[ht]
     \centering   
         \centering
         \includegraphics[width=1.0\textwidth]{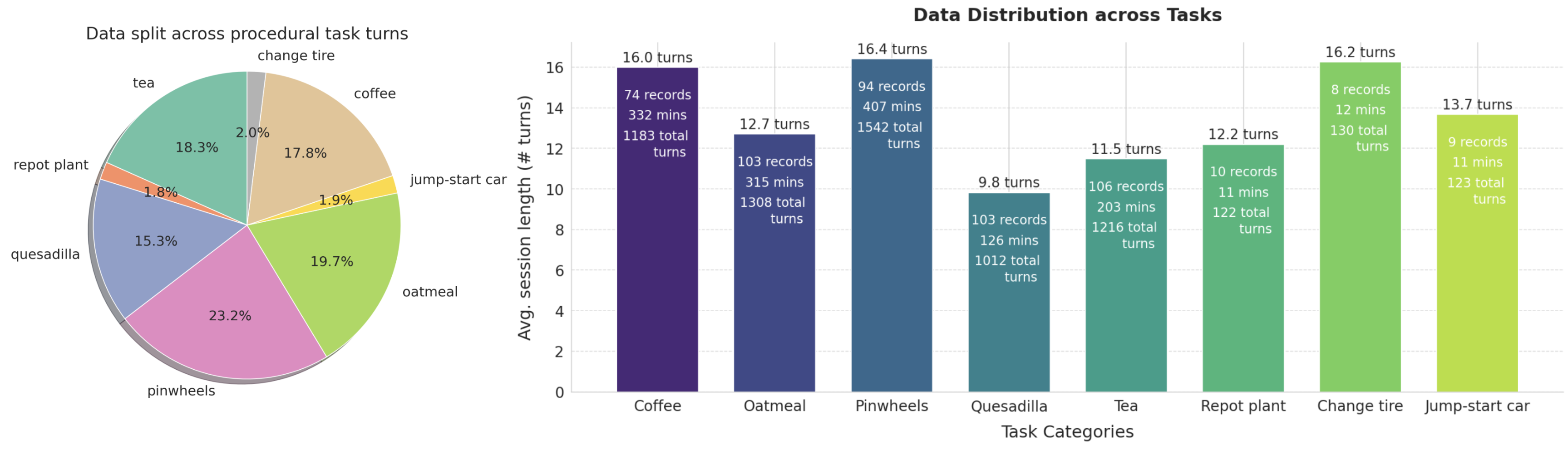}
         \caption{Distribution of HowToDIV dataset across various procedural tasks. Left: Proportion of user-expert turns for each task. Right: For each task - the average session length (measured as number of turns); the bars also show the total number of conversations, total video duration, and total number of turns.}
         \label{fig:task_split}
\end{figure*}

\section{HowToDIV Dataset}
\label{sec:dataset}

In this work, we introduce HowToDIV dataset comprising of dialogues, instructions and videosteps for two-person procedural task assistance scenarios. We describe the data curation in Sec.~\ref{sec:data_curation}. This is followed by annotations and statistics in Sec.~\ref{sec:data_statistics}, before moving to the data generation process in  Sec.~\ref{sec:method}.
\subsection{Data Curation}
\label{sec:data_curation}
We have built a dataset consisting of dialogues and video-clips between a novice user and expert guide for goal-oriented multi-step procedural tasks. HowToDIV consists of 507 recordings and 6636 dialogue turns with a total footage of 24 hours of videos representing user viewpoint. The dataset covers a diverse set of domains including cooking, mechanics, and planting, consisting of 9 tasks: making coffee, tea, oatmeal, pinwheels, quesadilla, jump starting a car, changing tires, and re-potting plants. Each session corresponds to a conversation initiated by a user inquiring about instructions for how to perform the specific task, followed by the expert describing the instructions. The conversation involves the user performing actions one step at a time, and requesting the next step once completed, along with any follow up clarification questions. The conversation ends with the user completing all the steps and finishing the desired task. Each user dialogue is accompanied with a video-clip corresponding to a recording from a user-worn wearable device capturing user's actions. The summarized instructions are also available for each conversation.

The dataset is is built on top of two publicly available instruction video datasets where a single person demonstrates the activities: Narrated Instruction Videos (NIV) ~\cite{niv} and Egocentric Procedural Error (EgoPER) Dataset ~\cite{egoper}.

\textbf{Narrated Instruction Videos}. The NIV~\cite{niv} dataset  consists of 150 YouTube videos comprising of an expert narrating the entire process for completing a particular task while demonstrating each action. It spans over five tasks covering 30 videos per task, with an average length of 2 minutes or 4,000 frames. These videos entail complex interactions between people and objects, covering both indoor and outdoor settings. The dataset provides English transcript for each video obtained via automatic speech recognition which are corrected for misspellings and punctuations by a human annotator. We manually select videos which have an egocentric viewpoint in order to align with our objective of extracting videos pertaining to user viewpoint.
\begin{table}[t]
    \centering
    \resizebox{0.45\textwidth}{!}{
    \begin{tabular}{cc}
     INSTRUCTION STEP & \multicolumn{1}{c}{USER ERROR} \\
     \midrule
     Place the tea bag in the mug & Dropped the tea bag on floor\\
     Roll the tortilla into a wrap & Folded the tortilla in half \\
     Fold the filter to create a semicircle & Tore the paper filter \\
    \bottomrule
     INSTRUCTION STEP & \multicolumn{1}{c}{MODIFICATION} \\
     \midrule
     Drizzle honey in bowl & Pour sugar instead of honey\\
     Stir using spoon & Stir using knife \\
     Slice tortilla using knife & Rip tortilla by hands \\
     Place tortilla on cutting board & Place tortilla on table \\
    
    \end{tabular}}
    \caption{Examples of (Top) User errors and (Bottom) Modifications and the corresponding instruction steps from HowToDIV}
    \label{tab:mistakes}
\end{table}

\textbf{Egocentric Procedural Error Dataset}. The EgoPER dataset ~\cite{egoper} represents a collection of 386 multimodal egocentric procedural task videos for cooking collected using Microsoft HoloLens2 through 11 participants. It includes multimodal data (RGB, depth, audio, gaze and hand) for five recipes: preparing pinwheels, quesadilla, oatmeal, coffee, and tea. The videos entail a participant demonstrating each step of the instruction however unlike NIV~\cite{niv}, audio or text narrations are not present. Instead the dataset provides metadata for the timestamp and fine-grained action description corresponding to each step of the recipe. The dataset not only consists of videos where the participant is performing correct actions, but also recordings where the participant makes errors and corrects them. Hence containing 213 normal, 173 erroneous videos for a total footage of 28 hours. The errors range across a variety of categories: step omissions, step additions, step modifications, step slips and step corrections. For HowToDIV dataset, we process all normal videos and only those erroneous video recordings that consist of step modifications and step corrections type mistakes. This is to help align our goal of conversations where a user makes a mistake in a step and is subsequently corrected by the expert in the following turn. 
\begin{figure}[ht]
     \centering   
     \includegraphics[width=0.45\textwidth]
     {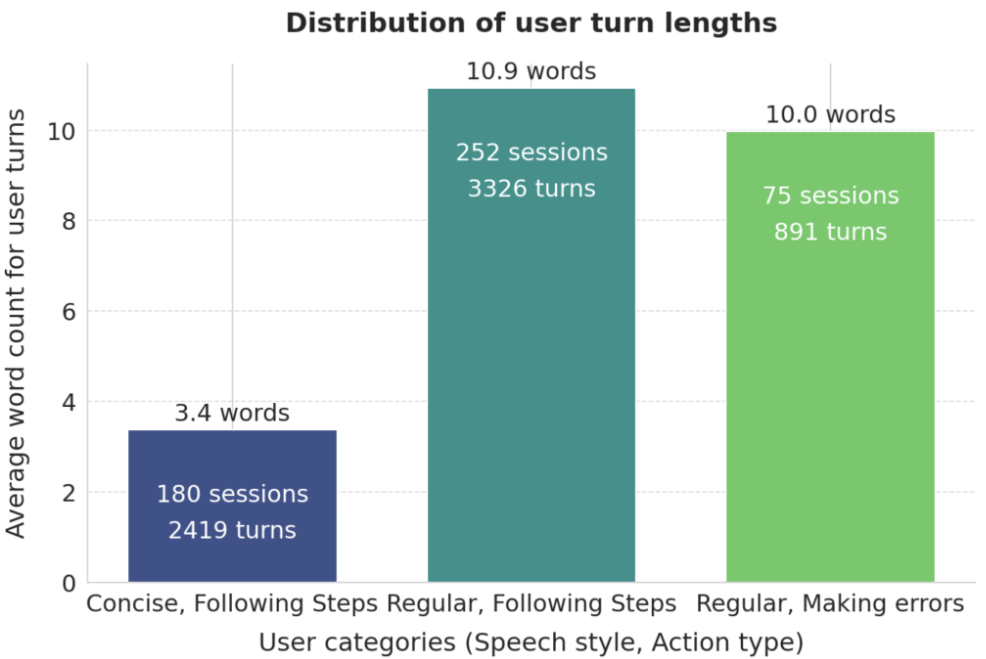}
     \caption{Distribution of user dialogue lengths across different speech \& activity categories. Category 1-Concise speech and user performs all steps accurately: the average is 3.4 words. Category 2-Regular speech and user performs steps accurately: the average is 10.9 words. Category 3:-Regular speech style however user makes errors: the average is 10 words. }
     \label{fig:howtodiv_user_turnlen}
\end{figure}

\begin{figure}[ht]
     \centering   
     \includegraphics[width=0.45\textwidth]
     {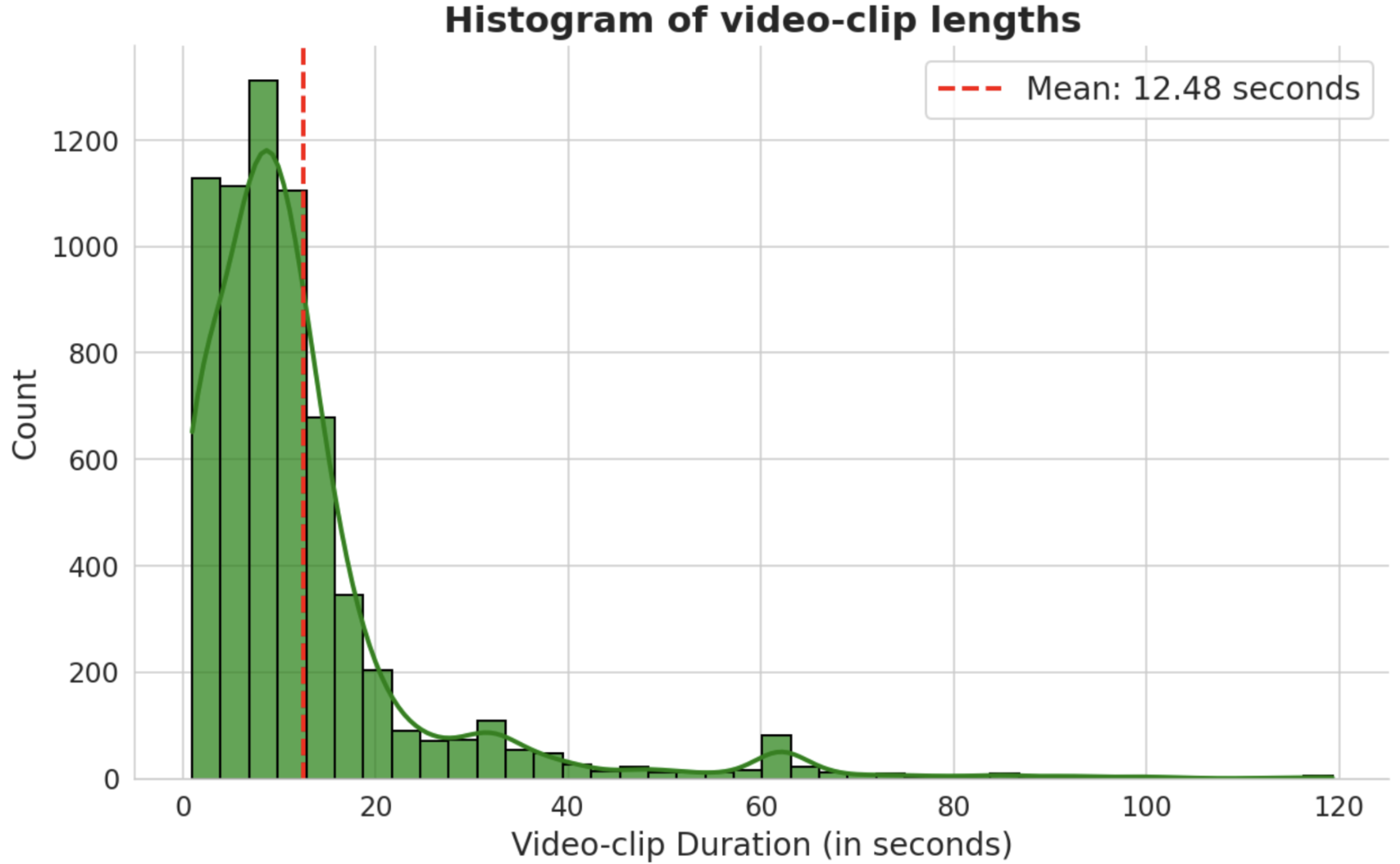}
     \caption{Histogram of user-turn video lengths, average being 12.48 seconds.}
     \label{fig:vid_len}
\end{figure}
\begin{table*}[t]
    \centering
    \resizebox{.95\textwidth}{!}{
    \begin{tabular}{c|c|cccc|cc|cc}
    \toprule   
     Dataset & Domain & Dialogues & Temporal-steps & Narrations & Instructions & Errors & Egocentric & Videos & Hours \\
     \midrule

     NIV~\cite{niv} & Varied & \xmark & \xmark & \cmark & \xmark & \xmark & \xmark & 6.66 & 150\\
     HowTo100M~\cite{howto100m} & Varied & \xmark & \xmark & \cmark & \xmark & \xmark & \xmark & 134.4k & 1.221M\\
     \hdashline\noalign{\vskip 0.5ex}
     EpicTent~\cite{epictent} & Tent Making & \xmark & \cmark & \xmark & \cmark & \cmark & \cmark & 7 & 24\\
     EgoPER~\cite{egoper} & Cooking & \xmark & \cmark & \xmark & \cmark & \cmark & \cmark & 28 & 386\\
     EGTEA~\cite{egtea} & Cooking & \xmark & \cmark & \xmark & \xmark & \xmark & \cmark & 28 & 10321\\
     Assembly101~\cite{assembly101} & Toy Assembly & \xmark & \cmark & \xmark & \cmark & \cmark & \cmark & 167 & 1425\\
     \hdashline\noalign{\vskip 0.5ex}
     EPICKitchen~\cite{epickitchen} & Cooking & \xmark & \cmark & \cmark & \xmark & \xmark & \cmark & 100 & 432\\
     COIN~\cite{coin} & Varied & \xmark & \cmark & \cmark & \xmark & \xmark & \xmark & 476 & 11827\\
     
     CrossTask~\cite{crosstask} & Varied & \xmark & \cmark & \cmark & \cmark & \xmark & \xmark & 376 & 4700\\
     YouCook2~\cite{youcook2} & Cooking & \xmark & \cmark & \cmark & \cmark & \xmark & \xmark & 176 & 2000\\
     \midrule
     
     HoloAssist~\cite{holoassist} & Assembly & \cmark & \cmark & \xmark & \cmark & \cmark & \cmark & 2221 & 166\\
     HowToDIV (Ours) & Varied & \cmark & \cmark & \xmark & \cmark & \cmark & \cmark & 576 & 24\\

    \bottomrule
    \end{tabular}}
    \caption{Comparison of HowToDIV with existing instructional video and dialogue datasets for procedural tasks.}
    \label{tab:dataset_comparison}
\end{table*}
\subsection{Annotations and Statistics}
\label{sec:data_statistics}
The distribution of the dataset across various tasks categories is depicted in Fig.~\ref{fig:task_split}. The average number of user-expert turns per session ranges from 9.8 turns for simpler tasks to 16.4 turns for complex ones, with an overall average being 13 turns. In terms of number of dialogues, the dataset is split approximately equally across different cooking tasks, while also including about 6\% data sessions in other domains. We show the number of turns, recording sessions and the video footage per task category in Fig.~\ref{fig:task_split}. In addition to categorizing across tasks, we also classify the data into multiple user speech-style and action-type categories as shown in Fig.~\ref{fig:howtodiv_user_turnlen}. HowToDIV consists of two distinct speech-styles for user dialogues: Concise dialogues and Regular dialogues. There are 180 concise speech sessions, and 327 regular speech sessions, with an average number of user-turn words being 3.4 words and 10.7 words respectively. Additionally, the regular speech sessions are further classified into two distinct action-type categories: Following-Steps and Making-Errors. There are 252 Following-Steps sessions where the novice accurately follows the instructions given by the expert; and 75 Making-Error sessions where the novice makes some kind of error during the task, and corrects it in the follow-up turn through expert help. We list some examples of user-errors and step modifications captured in our dataset in Tab. \ref{tab:mistakes}

Fig.~\ref{fig:howtodiv_expert_turnlen} also shows the distribution of expert turn lengths across all recordings, with an average of 19.3 words per expert turn. Since, each user turn is accompanied with a video-clip showing the actions performed by the user, the video clip durations vary from 1 second to over 2 minutes, shown in Fig.~\ref{fig:vid_len}. The average video-clip length for the dataset is 12.5 seconds. We compare HowToDIV dataset with existing instructional video datasets for procedural task in Tab. \ref{tab:dataset_comparison}, and note that HowToDIV is a first of its kind dataset consisting of instructions, dialogues and videos built in an extremely cost efficient manner.
\section{Method}
\label{sec:method}
\begin{figure*}[ht]
     \centering 
         \centering
         \includegraphics[width=1.0\textwidth]{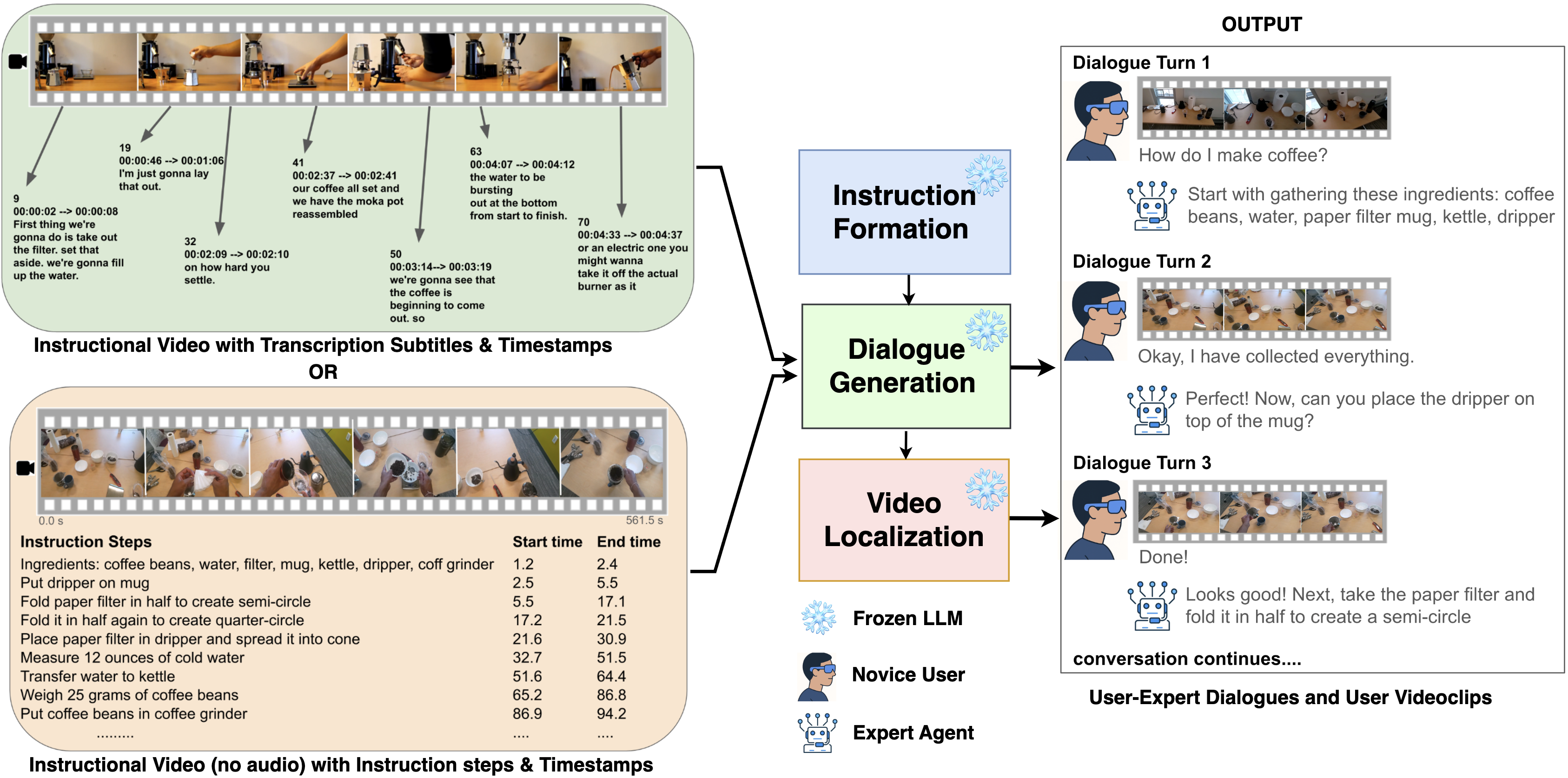}
         \caption{The input to our approach can be either: (Top-left) an Instructional video (with audio) and subtitles, OR (Bottom-left) an Instructional video (without audio) and step details. These are fed into our three stage approach resulting in a dialogue between user (equipped with a wearable, such as AI glasses) and expert, with each turn accompanied by user videos.}
         \label{fig:howtodiv_main_figure}
\end{figure*}
In this section, we present our method. It is a prompt-engineering based framework for transforming a monologue instructional video into two-person dialogue between a novice user and expert agent, utilizing a pre-trained language model with frozen weights. It leverages the idea that instructional videos captured from an egocentric viewpoint, have sufficient information to extract the task instructions comprehensively (i.e. recipe or the steps), along with fine-grained video-clips associated to each step. It further hinges on the idea that instructions can be utilized to synthesize a two-person conversation with an expert teaching the user how to accomplish the task. Our approach consists of three components: Instruction Formation, Dialogue Generation, and Video Localization, as shown in Fig.~\ref{fig:overview} and Fig.~\ref{fig:howtodiv_main_figure}. We briefly review the the Gemma 3 model in Sec.~\ref{sec:gemma3} and introduce each element of our pipeline in detail in Sec.~\ref{sec:instruction_formation}, ~\ref{sec:dialogue_generation} and ~\ref{sec:video_localization}
\subsection{Gemma 3}
\label{sec:gemma3}
Gemma3~\cite{gemma3} is an open-weight model family supporting text and image inputs via a SigLIP encoder. With a 128K-token context window, it handles long documents and image inputs. Instruction-tuned variants deliver strong multimodal, multilingual performance while remaining lightweight for edge use.
\subsection{Problem Setting}
\label{sec:problem_formulation}
The problem setup involves a user wearing an AR device, such as AI glasses, embedded with sensors such as a camera, speaker and microphone. Through this device, the user communicates with an integrated AI agent that can listen to user queries as well as perceive user's surrounding via real-time audio-video streaming. The goal of the agent is to assist the user perform a procedural task of their choice by providing instructions one step at a time. In this setup, the input to the agent consists of natural language queries from the user (e.g., "Can you teach me how to make pesto pasta" or "What do I do next?"), along with short egocentric video clips capturing the user's surroundings. For the output, the agent is expected to respond with the next actionable instruction (e.g., "First, gather ingredients like fusilli, basil, vegetables,..." or "Now add 2 cups of water to the pan.").

To address this problem, a dataset of multi-turn dialogues, where each turn comprises of a user query, an egocentric user videoclip and the expert's response, is needed. As shown in Fig.~\ref{fig:howtodiv_main_figure}, we propose constructing such a dataset by leveraging egocentric instructional videos, which typically appear in two flavors:

\begin{enumerate}
    \item A video (with audio) consisting of an expert narrating the instructions while demonstrating each step, consisting of $N$ image frames denoted as $V = (I_1, I_2, \dots, I_N)$ accompanied with subtitles $S$ obtained via transcription, $S = (S_1, S_2, \dots, S_M)$ comprising of $M$ entries. Each subtitle entry $S_j = (s_j, t_{js}, t_{je})$ comprises of the narration text $s_j$, start timestamp $t_{js}$ and end timestamp $t_{js}$ from the video.
    \item A video (without audio) consisting of an expert performing the steps for the task, denoted as $V = (I_1, I_2, \dots, I_N)$, along with annotations for steps represented as $Q_j = (q_j, t_{js}, t_{je})$ comprising of the step-description text $q_j$, start timestamp $t_{js}$ and end timestamp $t_{js}$ from the video.
\end{enumerate}

We formulate the task of dialogue generation as yielding a multi-turn conversation $C = {(c_1, c_2, \dots, c_P)}$ of $P$ turns where each turn $c_i = {(u_i, e_i, t_{is}, t_{ie})}$ consists of user query $u_i$, expert (or model) response $e_i$, start timestamp $t_{is}$ and end timestamp $t_{ie}$ of the video-clip. This dataset format is dictated by how modern-day LLM’s are prompted for training and inference.

\subsection{Instruction Formation}
\label{sec:instruction_formation}
The first stage of our pipeline produces a comprehensive and itemized set of task instructions. Each step is defined to be optimally atomic, corresponding to a single consumable action. Given the diversity of instructional video metadata - ranging from narrations to step labels and fine-grained action annotations - we adopt tailored extraction strategies. For instructional videos with audio narrations, for e.g., NIV~\cite{niv}, we obtain subtitle transcripts from existing annotations or via an automatic speech recognition module. These transcripts are then fed into a large language model, which is prompted to reason out the key steps and augment the extract latent information with it’s inherent knowledge. For videos with fine-grained action labels, such as EgoPER~\cite{egoper}, we perform lightweight post-processing to merge duplicates and enhance label descriptions. For datasets such as EGTEA~\cite{egtea}, we cluster semantically similar actions and filter out generic, non-essential steps (e.g., open drawer, close container) to yield a concise yet complete instruction set.
\subsection{Dialogue Generation}
\label{sec:dialogue_generation}
Once the instruction steps are obtained in an itemized list format, we perform a second LLM inference to convert them into cumulative question–answer pairs, simulating a dialogue between a novice agent (without access to the instructions) and an expert agent (with full access). The prompt is tuned to request each atomic step being mapped to a single dialogue turn, with both concise and natural-speech variants generated separately. To model error scenarios, we leverage datasets such as EgoPER~\cite{egoper}, which annotate steps as correct or  erroneous along with their corrective actions. For this, the instruction list is modified to mark the corrective steps with special tokens, enabling the LLM to produce expert dialogues that identify and address mistakes. Additionally, procedural caveats, special notes, and nuanced constraints from narrations are converted into clarification questions, further enriching the interaction with fine-grained domain knowledge.
\subsection{Video Localization}
\label{sec:video_localization}
The final stage involves temporally localizing each atomic instruction step within the original instructional video, to identify the corresponding video segment. This is meant to be repurposed as the user’s egocentric visual stream for performing that step. The extraction method depends on the source dataset. For datasets with fine-grained step annotations (e.g., EgoPER, EGTEA), we directly use the provided start and end timestamps, either for the individual step or by aggregating timestamps for clustered steps, depending on how the instructions were formed. For YouTube-style instructional videos containing only narrations, we leverage subtitle timestamps generated as part of ASR output, to estimate segment boundaries for each step. The action timestamps are obtained during the instruction extraction stage in the same LLM call. Incorporating more advanced temporal action localization techniques is left for future work.
\section{Experiments}
\label{sec:experiment}
\subsection{Implementation Details}

We use Gemma-3 model~\cite{gemma3} as the base LLM for generating instructions, dialogues and video labels for HowToDIV dataset, specifically the most powerful instruction-tuned variant consisting of 27B parameters.
We utilize FSDP Sharding over TPU v3 with 8 compute nodes and 4 Dragonfish chips for all experiments. 
We randomly split the 507 sessions into train, validation and test sets following a ratio of 70\%, 10\% and 20\%, on a per-task, per user-category basis, resulting in 355 sessions for training, 44 sessions for validation and 108 sessions for testing. All results are reported on the test-split.

\textbf{Diversity Matters.} To ensure sufficient diversity in the generated dialogues, we use Random Sampling with a temperature score of 1.5. This enables varying user-response and expert=instruction styles as illustrated in Fig.~\ref{fig:howtodiv_main_figure}. We also generate two flavors of conversations for all normal sessions from EgoPER dataset corresponding to regular user speech and concise user speech.

\textbf{Inference - Expert Response Generation.} To establish the baseline performance on HowToDIV, we use the 4B variant of Gemma-3 model in the text input setup for generating expert responses, and defer the multi-modal input setup for future work. For the prompt we experiment with two different templates. In the first setup 'Hint + Steps', we include the full instructions for the given task in the session in system prompt, to ease the model for generating responses. In this setup, the model is expected to reason well, track user state accurately, and output the correct action response. In the second, more realistic and challenging, setup 'Hint-only', we do not provide the instructions, and instead rely on the internal domain knowledge learned by the model. 
\begin{figure}[ht]
     \centering   
     \includegraphics[width=0.45\textwidth]
     {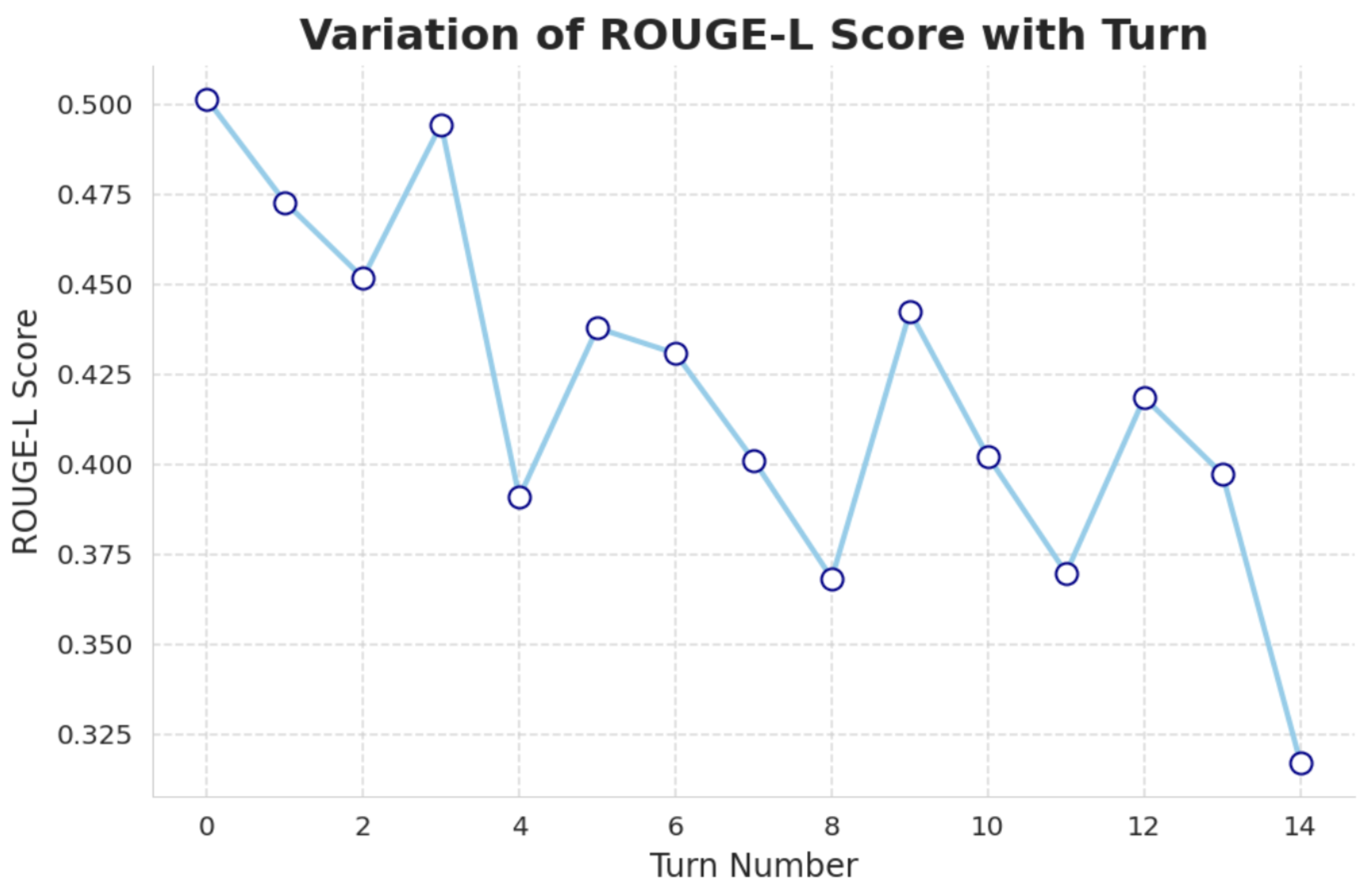}
     \caption{Variation of average ROUGE-L score with increasing dialogue turns for HowToDIV conversations}
     \label{fig:bleu_turnwise}
\end{figure}

\textbf{Metrics.} Following prior works ~\cite{ji2019invariant,caron2018deep,cho2021picie,hamilton2022unsupervised,zhou2022extract,shin2022reco}, we use three metrics to measure the expert response generation performance at inference. 

\begin{figure}[ht]
     \centering   
     \includegraphics[width=0.51\textwidth]
     {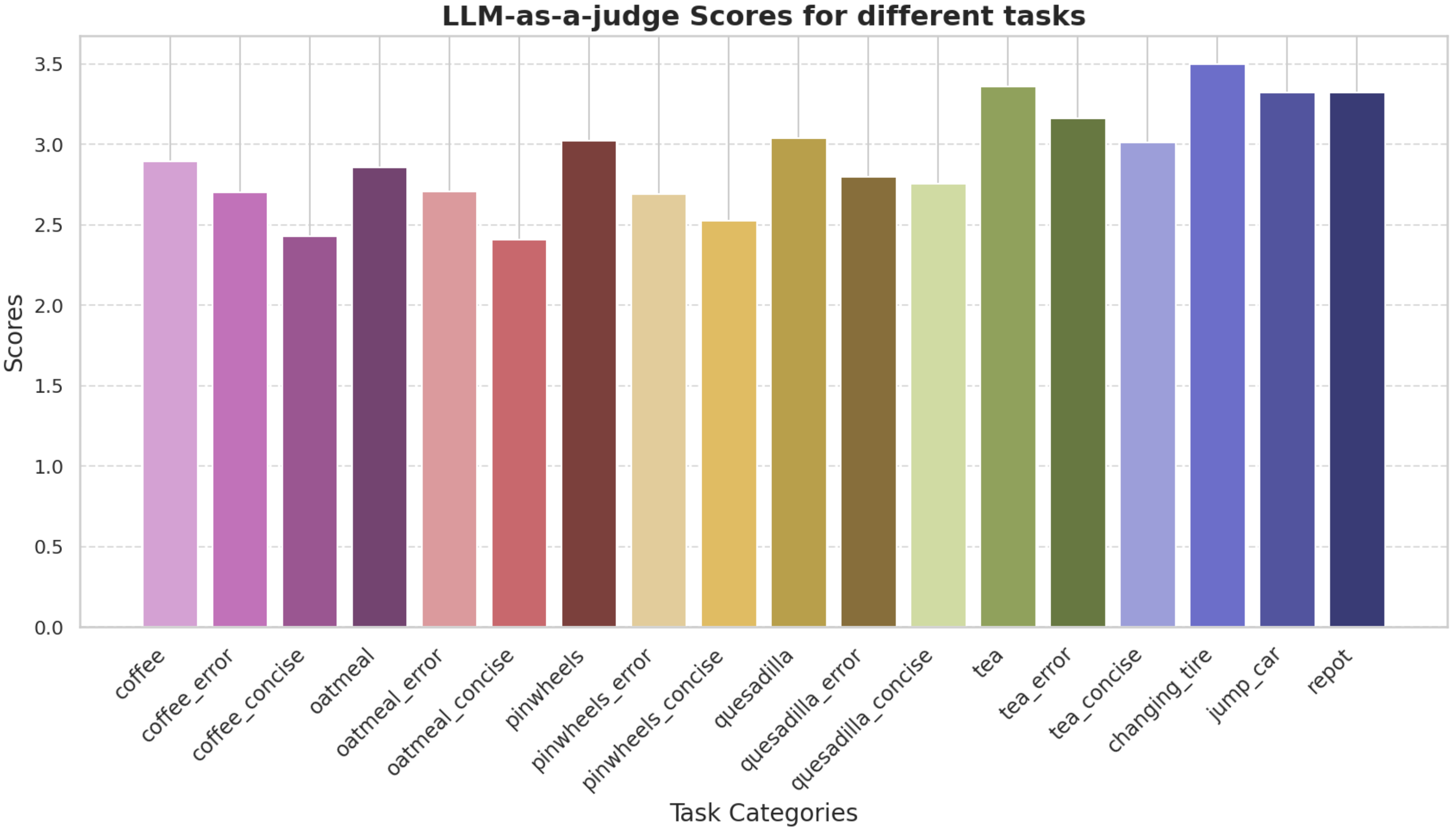}
     \caption{Variation of LLM as a judge scores across different tasks, speech-styles and action-categories.}
     \label{fig:llmj_taskwise}
\end{figure}
\noindent\textbf{LLM as a Judge.} LLM-as-a-judge ~\cite{llmasjudge},~\cite{mllmasjudge} uses pre-trained models, LLM's, to evaluate responses automatically, providing human-like evaluation. It first defines an evaluation prompt based on the setup along with a trace or a dataset entry, and asks the model to score and reason the output. It has been considered a scalable, accurate, and reliable alternative to costly human ratings. In our evaluation setup, we prompt a Gemma-3, 12B model to output the reasoning along with a score between 1 and 5.

\noindent\textbf{ROUGE Score.} The ROUGE~\cite{rouge} score (Recall-Oriented Understudy for Gisting Evaluation), typically used in text summarization and machine translation, measures the overlap between generated text and reference text. It ranges between 0 and 1, with higher scores indicating higher similarity. ROUGE-N evaluates the overlap of n-grams, focusing on the recall aspect i.e. how much of the reference text is captured in the generated text.

\noindent\textbf{BLEU Score.} The BLEU~\cite{bleu} score (Bilingual Evaluation Understudy) is typically used to evaluate the quality of machine-translated text by comparing with a reference translation. It ranges from 0 to 1, with 1 indicating a perfect match to the reference, calculating the precision of n-grams, i.e. the proportion of n-grams in the machine translation that also appear in the reference translations. 
\begin{figure}[ht]
     \centering   
     \includegraphics[width=0.45\textwidth]
     {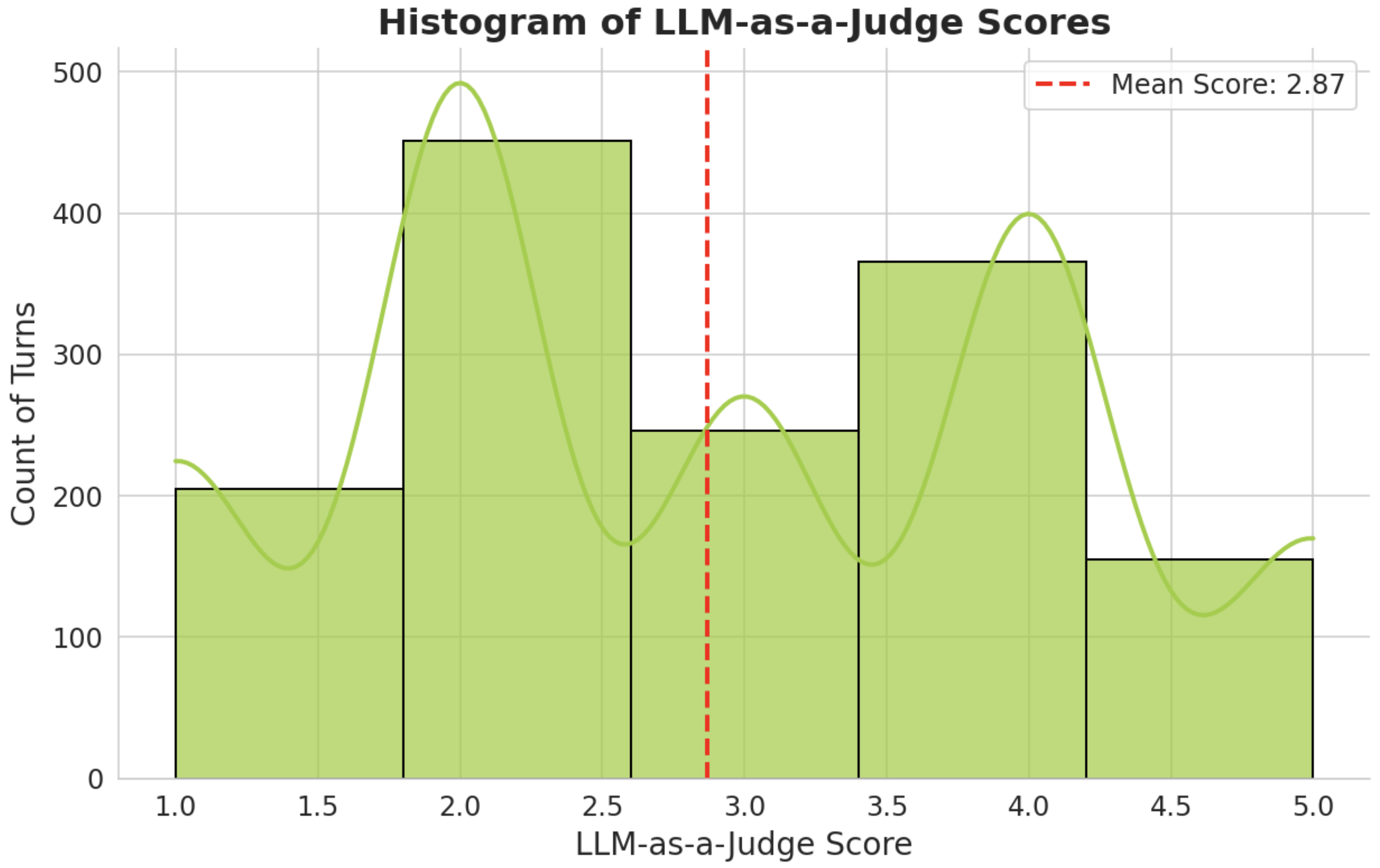}
     \caption{Histogram of LLM-as-a-Judge scores on HowToDIV corresponding to Hint-only prompt setting.}
     \label{fig:llm_aaj_scores}
\end{figure}

\subsection{Results and Observations}
\label{sec:main_results}
\begin{table}[t]
    \centering
    \resizebox{0.5\textwidth}{!}{
    \begin{tabular}{c|ccccc}
    \toprule   
     Method & \multicolumn{5}{c}{Metrics} \\
      & BLEU  & LLM-Judge & ROUGE-1 & ROUGE-2 & ROUGE-L \\
     \midrule
    Gemma-3 (Hint only) & 0.321 & 2.870 & 0.325 & 0.125 & 0.270 \\
    Gemma-3 (Hint + Steps) & 0.457 & 4.101 & 0.489 & 0.268 & 0.429 \\
    \bottomrule
    \end{tabular}}
    \caption{Evaluation on HowToDIV for Gemma-3 responses, reported for Hint-only and Hint + Steps prompt configurations. 12B Gemma-3 variant is used for LLM-as-a-Judge metric while 4B version is used for inference responses}
    \label{tab:howtodiv_metrics}
\end{table}
\begin{table}[t]
    \centering
    \resizebox{0.5\textwidth}{!}{
    \begin{tabular}{c|ccccc}
    \toprule   
     User category & \multicolumn{5}{c}{Metrics} \\
      & BLEU  & LLM-Judge & ROUGE-1 & ROUGE-2 & ROUGE-L \\
     \midrule
    Concise-Follow & 0.307 & 2.616 & 0.309 & 0.116 & 0.258 \\
    Regular-Follow & 0.335 & 3.066 & 0.342 & 0.135 & 0.282\\
    Regular-Error & 0.304  & 2.816 & 0.304 & 0.109 & 0.257 \\
    \bottomrule
    \end{tabular}}
    \caption{Evaluation results across various user-speech style and action-types on HowToDIV, corresponding to inference with Gemma-3, for Hint-only prompt setup.}
    \label{tab:usertype_metrics}
\end{table}
\begin{figure}[ht]
     \centering   
     \includegraphics[width=0.49\textwidth]
     {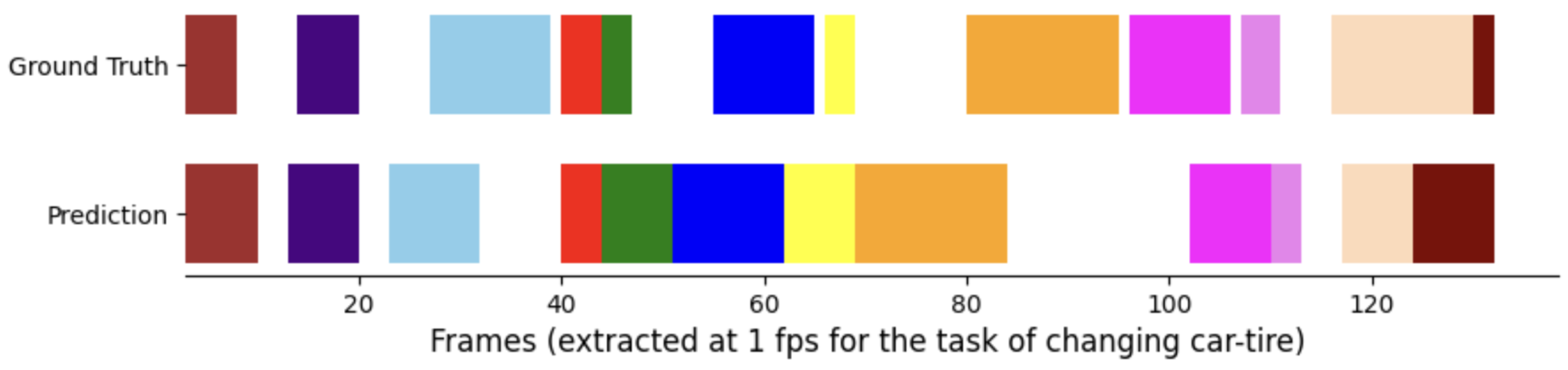}
     \caption{Visualization of temporal action localization - Ground truth and Predicted steps. The video is associated with the task of changing car tire, each color represents a different step.}
     \label{fig:action_segmentation_car}
\end{figure}


We compare the performance of Gemma-3 model on our two distinct prompting setups: Hint-Only where we prompt the model with the hint to output responses one step at a time; and Hint + Steps, which includes the instructional steps for the given task appended with the base prompt. In Tab.~\ref{tab:howtodiv_metrics}, we observe that the performance of the Hint+Steps setup is significantly better than that of the Hints-only across all metrics. This is expected since without the task instructions, the LLM relies on inherent model knowledge which often suffers from hallucinations and domain deficiencies. We observe that without finetuning, Gemma-3, 4B model yields a 2.87 LLM-as-a-Judge score, 0.321 BLEU and 0.27 ROUGE-L score. In Fig.~\ref{fig:llm_aaj_scores} we see the median LLM-as-a-Judge score to be 3.0, on a 1 to 5 scale, for the Hint-only setup. We also observe in Fig.~\ref{fig:bleu_turnwise} that model performance decreases as the conversation progresses, shown as dropping curve with increasing turn numbers. A similar trend of reducing accuracy with turn numbers is observed across other metrics as well. Across the different user-speech and action-type categories, we observe that for most tasks, the performance of regular-speech is superior to concise-speech recordings as seen in Tab.~\ref{tab:usertype_metrics} and Fig.~\ref{fig:llmj_taskwise}, as observed for all metrics. Moreover, the performance of conversations where the user follows-steps is also significantly higher compared to those where the user makes errors. This trend is observed across most tasks as seen in Fig.~\ref{fig:llmj_taskwise}. Between ROUGE-1 and ROUGE-2 metrics, we observe that ROUGE-1 scores corresponding to monograms are higher than ROUGE-2 scores corresponding to bi-grams as expected.

We also visualize the performance of action segmentation in Fig. ~\ref{fig:action_segmentation_car} comparing a human-annotated ground truth with time label predictions from our algorithm, for a recording from NIV dataset, and observe an IOU score of 0.392, precision of 54.84\% and accuracy of 57.95\%.

\section{Conclusion}
\label{sec:conclusion}
In this paper, we presented a novel method for transforming monologue instructional videos into two-person, user-expert dialogue sessions grounded in fine-grained video segments and instruction steps. Our approach leverages the language capabilities of large language models to generate rich, multi-turn dialogues and synchronized user viewpoint video clips without manual annotation or collection. Using this method, we introduced HowToDIV, a large-scale dataset consisting of 507 conversations and 6636 question answer or dialogue turns across diverse real-world tasks in cooking, mechanics, and planting. HowToDIV provides a scalable and cost-efficient solution to the scarcity of procedural task-assistance dialogue datasets, opening new avenues for training and evaluating AI agents in realistic scenarios. We also benchmark HowToDIV using Gemma3-4B model, establishing strong baseline metrics for future research. Our work lays the foundation for building conversational agents capable of guiding users through complex multi-step procedures in real time.

\newpage
{\small
\bibliographystyle{ieee_fullname}
\bibliography{egbib}
}

\section{Appendix}
\label{sec:appendix}
\subsection{Gemma3 Model}
\label{sec:gemma3_v2}
Gemma3-4B \cite{gemma3} is one of four open-weight models in Google’s Gemma3 family, comprising approximately four billion parameters. It supports both text and image inputs through a custom SigLIP vision encoder combined with a Pan-and-Scan preprocessing technique, which converts visual data into compact 256-token “soft” representations. The model features a 128K-token context window—16× larger than in previous Gemma versions—enabling efficient processing of extended multi-page text, large single documents, or hundreds of images in a single prompt. Its architecture alternates five local-attention layers (1024-token sliding windows) with a global-attention layer, balancing short-range context modeling with long-range dependency tracking, while optimizing KV-cache memory usage. The instruction-tuned variant, Gemma-3-4B-IT, achieves benchmark performance comparable to, and in some cases exceeding, that of the much larger Gemma2-27B model. With strong multimodal capabilities, multilingual coverage across 140+ languages, and a lightweight design suitable for single-GPU or edge deployment, Gemma3 4B offers a compelling solution for resource-constrained environments.

\begin{figure}[ht]
     \centering   
     \includegraphics[width=0.45\textwidth]
     {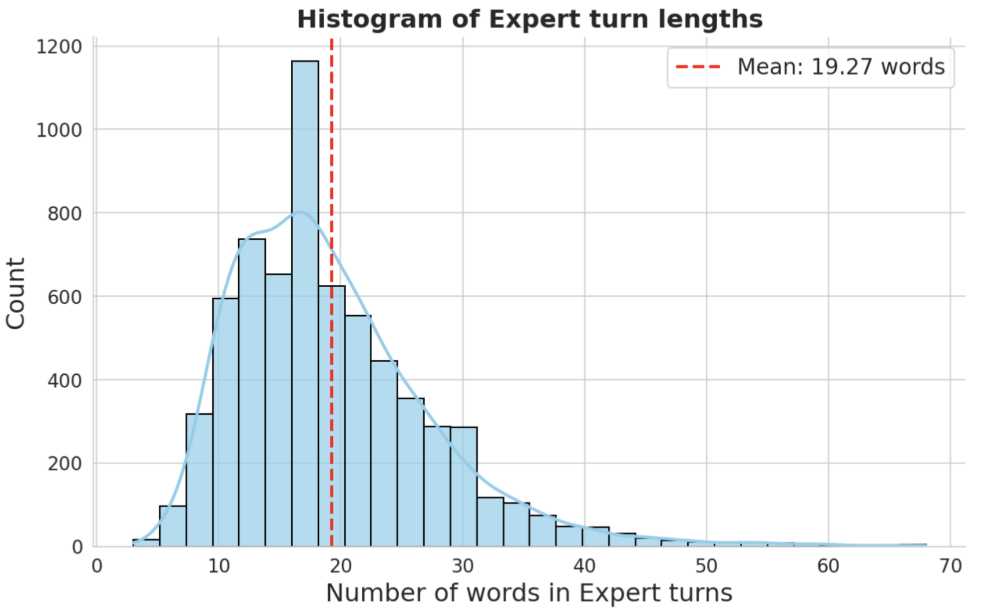}
     \caption{Histogram of Expert dialogue lengths, the average being 19.3 words.}
     \label{fig:howtodiv_expert_turnlen}
\end{figure}
\subsection{Expert Turn Statistics}

We visualize the histogram for expert turn lengths in Fig.~\ref{fig:howtodiv_expert_turnlen} and observe longer dialogues compared to user turns, average being 19.3 words.

\section{Limitations and Future Work}
\label{sec:limitations_future_work}
In this paper, we present our approach, contribute the HowToDIV dataset and benchmark its performance on the problem of procedural-task assistance. Our multimodal  dataset (text and video) lends a natural fit to other language-vision tasks as well. 1) This includes fine-grained action recognition~\cite{fgactionrecognition}, instruction guided video generation~\cite{fgvideogeneration} and video retrieval such as for improving VideoCLIP~\cite{videoclip}. Owing to the availability of large number of videos for fine-grained steps, training with HowToDIV can improve performance significantly. 2) While we lay a strong foundation with HowToDIV, numerous extensions are possible to scale it across multiple dimensions and domains. Using our approach, video datasets with temporal-steps labels such as EpicKitchen~\cite{epickitchen}, CrossTask~\cite{crosstask}, YouCook2~\cite{youcook2}, EpicTent~\cite{epictent}, EGTEA~\cite{egtea} among the others in Table~\ref{tab:dataset_comparison} can be transformed into conversations to expand HowToDIV 3) We consider proactivity, out-of-order steps, user errors, and corrections to be crucial aspects of task guidance, and encourage dataset extensions that address these dimensions. 4) Recent image conditioned video generation methods such as FramePack~\cite{framepack} have opened opportunities to generate videos with consistent backgrounds and task setups, enabling the simulation of user mistakes and corrections as a promising direction for video–dialogue data augmentation. 5) For the specific task of procedural-task assistance dialogues, we see LLM finetuning on HowToDIV a natural progression to our research, in addition to continued efforts to explore more robust evaluation protocols. Our approach offers a low-cost yet powerful methodology that leverages readily available resources to generate large-scale dialogue datasets. We hope it inspires the community to further pursue research in this direction.

\clearpage
\newpage

\end{document}